\documentclass{article}
\usepackage{microtype}
\usepackage{graphicx}
\usepackage{subfigure}
\usepackage{amssymb}
\usepackage{amsmath}
\usepackage[hyphens]{url}
\usepackage{bbding}
\usepackage{hyperref}
\usepackage{booktabs} 

\usepackage{hyperref}

\usepackage[accepted]{icml2021}

\icmltitlerunning{Multi-Modal Deep Learning Approach for Post-Hurricane Building Damage Assessment}

\begin{document}

\twocolumn[
\icmltitle{Post-hurricane building damage assessment using street-view imagery and structured data: A multi-modal deep learning approach}




\begin{icmlauthorlist}
\icmlauthor{Zhuoqun Xue}{uf}
\icmlauthor{Xiaojian Zhang}{uf}
\icmlauthor{David O. Prevatt}{uf}
\icmlauthor{Jennifer Bridge}{uf}
\icmlauthor{Susu Xu}{jhu}
\icmlauthor{Xilei Zhao}{uf}
\end{icmlauthorlist}

\icmlaffiliation{uf}{Department of Civil and Coastal Engineering, University of Florida, Gainesville, Florida, USA}
\icmlaffiliation{jhu}{Department of Civil and Systems Engineering, Johns Hopkins University, Baltimore, Maryland, USA}

\icmlcorrespondingauthor{Xilei Zhao}{xilei.zhao@essie.ufl.edu}

\icmlkeywords{Machine Learning, ICML}

\vskip 0.3in
]



\printAffiliationsAndNotice{}  

\begin{abstract}
Accurately assessing building damage is critical for disaster response and recovery. 
However, many existing models for detecting building damage have poor prediction accuracy due to their limited capabilities of identifying detailed, comprehensive structural and/or non-structural damage from the street-view image. 
Additionally, these models mainly rely on the imagery data for damage classification, failing to account for other critical information, such as wind speed, building characteristics, evacuation zones, and distance of the building to the hurricane track. 
To address these limitations, in this study, we propose a novel multi-modal (i.e., imagery and structured data) approach for post-hurricane building damage classification, named the Multi-Modal Swin Transformer (MMST). 
We empirically train and evaluate the proposed MMST using data collected from the 2022 Hurricane Ian in Florida, USA. 
Results show that MMST outperforms all selected state-of-the-art benchmark models and can achieve an accuracy of 92.67\%, which are 7.71\% improvement in accuracy compared to Visual Geometry Group 16 (VGG-16). In addition to the street-view imagery data, building value, building age, and wind speed are the most important predictors for damage level classification. The proposed MMST can be deployed to assist in rapid damage assessment and guide reconnaissance efforts in future hurricanes.
\end{abstract}


\section{Introduction}

Due to climate change and population growth in coastal areas, hurricanes are causing more damage and loss to the U.S., e.g., the 2021 Hurricane Ian \cite{ian_report}. Accurately and timely assessing hurricane-induced building damage is critical for emergency response and community recovery \cite{wang2023causality}. Despite much research devoted to this topic, significant research gaps remain.

First, emerging post-hurricane building damage assessment methods typically relied on satellite imagery data \cite{calton2022using,lin2021building, gupta2019xbd, chen2021using, duarte2018satellite, wang2024scalable}. However, such data often struggle with weather and environmental conditions and fail to reveal critical external damages like wall coverings or structural integrity \cite{mccormack2011capabilities,lee2005application,wang2020deep}. On the other hand, street-view imagery data, which captures almost the full view of a building, can largely overcome these limitations \cite{chen2022deep}. Yet, how to use street-view imagery data for accurate damage assessment remains largely unsolved \cite{berezina2022hurricane, zhai2020damage}.

Second, most of the damage assessment models proposed in existing studies used Convolutional Neural Network (CNN) as the backbone model \cite{hong2022classification, xiong2021multiple, ci2019assessment, seydi2022bdd, zhai2020damage}; however, CNN has several major limitations. Specifically, CNN first extracts features through fixed-sized convolutional kernels. This approach makes CNN susceptible to noise in images (such as debris and trees), leading to reduced prediction accuracy \cite{sriwong2021study}. In addition, the architecture of CNN focuses on learning building damage patterns through local information, resulting in its incapability of grasping the overall damage level to a building. Recent studies have found that attention-based Transformers can effectively distinguish between important information and noise in images as well as form a comprehensive understanding of all the important damage information contained in an image \cite{kaur2023large,chen2022dual}, showing great potential in enhancing model accuracy. However, research and application of Transformer in post-hurricane building damage assessment are still limited \cite{guo2022transformer, mirzapour2023capturing}.

\begin{figure*}[h]
  \centering
  \includegraphics[width=0.8\linewidth]{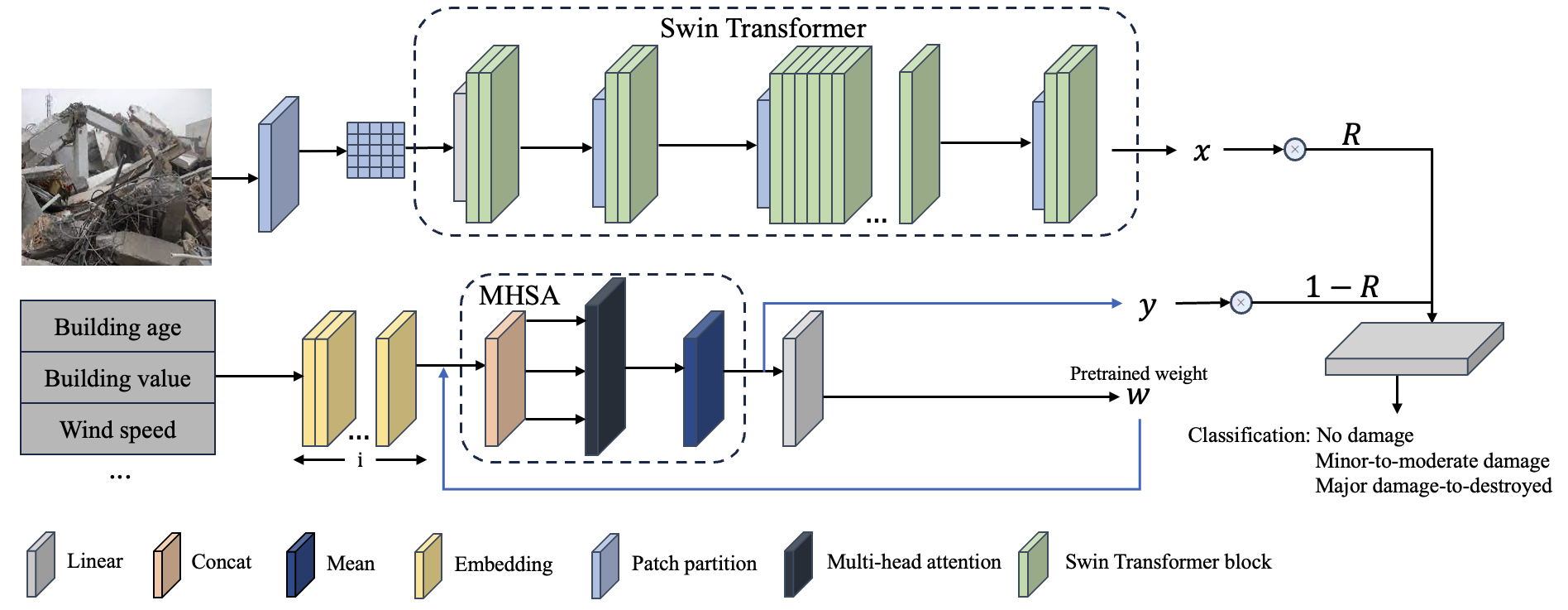}
  \caption{MMST architecture.}
  \label{fig:Figure 1}
\end{figure*}

Third, most previous studies only adopted a single mode of data, i.e., imagery data, for assessing building damage levels \cite{zhai2020damage, cheng2021deep}. However, relying on a single data modality is probably insufficient for a complete damage depiction, which may limit the model's prediction performance. For example, models depending only on images can only achieve the damage prediction accuracy of approximately 70\% or even less \cite{zhai2020damage}. Moreover, if the single data source is flawed or incomplete, the predictions generated by the model could be skewed \cite{cheng2021deep}. For example, buildings could be (partially) blocked by vegetation after hurricanes. Accordingly, models can only assess the damage levels based on the visible parts of an image, which may only partially capture the actual extent of damage. According to our domain knowledge, in addition to imagery data, other multi-modal data, such as wind speed, building age, and hurricane path, were found to be influential to the building damage levels \cite{cheng2021deep, fronstin1994determinants, egnew2018linking, xian2015storm, jain2009statistical}. We believe these factors can provide additional insights (which may not be captured by imagery data) into the building damage assessment process. Overlooking such domain-knowledge-related factors may limit the damage prediction accuracy and subsequently affect the reliability of downstream applications \cite{zhang2022machine, borghesi2020improving}. However, models that can account for multiple data modalities when assessing building damages during disasters are nearly nonexistent \cite {hao2020leveraging}. 

To overcome these limitations, we propose a novel methodology called Multi-Modal Swin Transformer (MMST) to classify the building damage levels post-hurricane rapidly. Specifically, MMST consists of three components. First, we utilized Swin Transformer (Swin-S) \cite{liu2021swin} as the backbone network of the image feature extractor to process the imagery modal. The main reason to choose it is that its (shifting) window self-attention mechanism efficiently handles features at different scales and captures global relationships, thus enhancing its capabilities in image classification tasks. Second, the structured data feature extractor, which utilizes multi-head self-attention (MHSA) \cite{vaswani2017attention}, weighs the importance of multiple sets of structured data input to the MMST and generates representations. This is used to enhance the model's performance \cite{huang2021makes}. Third, the multi-modal data fusion module integrates image and structured data representations utilizing an adjustable fusion ratio for the final building damage classification. We test the proposed MMST using an empirical multi-modal dataset from the 2022 Hurricane Ian. The dataset contains the street-view imagery data for buildings affected by the hurricane and hurricane-related structured data inspired by previous studies such as wind map, the proximity to hurricane track, and specific building characteristics such as building value, age, and wind speed \cite{hao2020leveraging,chen2022rapid,wang2023causality,alam2020descriptive,imran2020using}. The results show that compared to the performance of the state-of-the-art benchmark models, MMST has the best prediction accuracy (92.67\%). Once deployed, the model can help emergency management agencies to facilitate rapid emergency response and guide reconnaissance efforts.

This study contributes three-fold to the literature:

\begin{itemize}
\vspace{-0.5cm}
    \item We use a novel attention-based transfer learning technique, i.e., Swin Transformer, to extract image features from post-hurricane building damage imagery. It significantly outperforms the commonly used CNN-based models in terms of both prediction performance and damage identification.
\vspace{-0.4cm}
    \item We propose a multi-modal deep learning model, i.e., MMST, for post-hurricane building damage assessment. In addition to imagery data, MMST can also incorporate domain-knowledge-related structured data such as building characteristics and wind speed.
\vspace{-0.4cm}
    \item MMST achieves Matthews correlation coefficient, sample-weighted F1 score, and accuracy of 0.7404, 0.9386, and 92.67\%, respectively. Compared to VGG-16, The improvement are 56.70\%, 5.03\%, and 7.71\%, respectively.
\end{itemize}

\section{Methodology}
\begin{figure*}[h]
  \centering
  \includegraphics[width=0.7\linewidth]{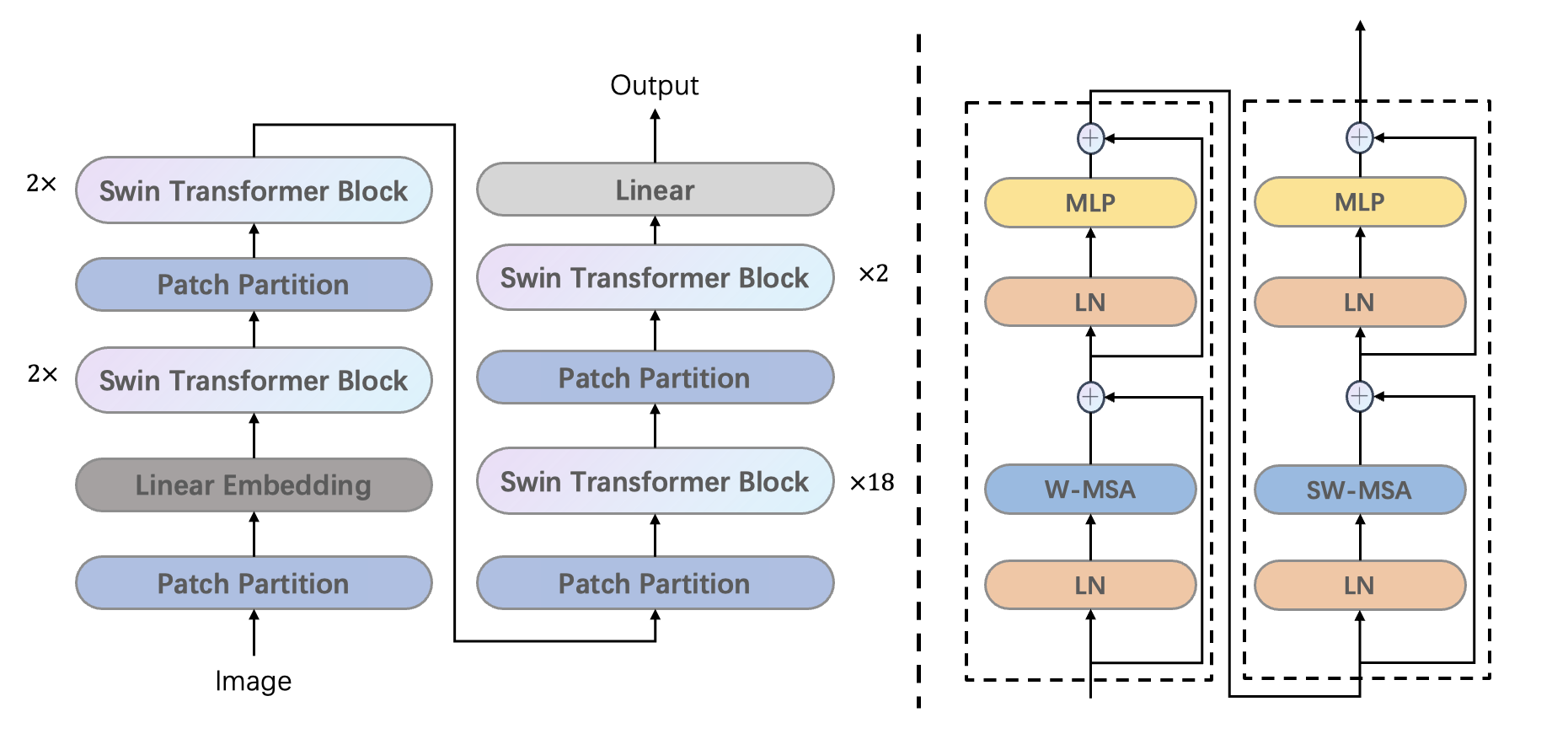}
  \vspace{-0.5cm}
  \caption{The Swin Transformer architecture (Swin-S). On the right is the structure of Swin Transformer block.}
  \label{fig:Figure 2}
\end{figure*}
As shown in Figure \ref{fig:Figure 1}, this study proposes a novel multi-modal classification model for post-hurricane building damage. MMST consists of three parts: image feature extractor, structured data feature extractor, and multi-modal fusion. First, we applied Swin Transformer as the backbone of the image feature extractor to realize the efficient feature extraction function based on shifted windows. Second, using the multi-head self-attention (MHSA) module in the structured data feature extraction process can enable the model to capture the complex relationships between structured data. Finally, through multi-modal fusion, image features and structured data features can simultaneously participate in the classification decision according to different fusion ratios.

\subsection{Image Feature Extractor}
The Transformer was first proposed and designed by \citet{vaswani2017attention}, and its main components are Multi-Head Attention (MHA) and Feed-Forward Networks. Inspired by the original Transformer structure, the researchers believe its efficient architecture for processing continuous sequences can be well applied in computer vision. \citet{liu2021swin} proposed a powerful vision Transformer named Swin Transformer. Swin Transformer has become a general-purpose backbone in computer vision due to its efficient self-attentive computation and the design of hierarchical feature maps.

The structure of the Swin Transformer is shown in Figure \ref{fig:Figure 2}. First, the input image will be divided into multiple non-overlapping patches and linearly embedded to extract initial features. Then, these features will be organized into non-overlapping windows and passed into the Swin Transformer block. The Swin Transformer block contains two sub-modules: the window multi-head self-attention module (W-MSA) and the shifting window multi-head self-attention module (SW-MSA). W-MSA enables the model to capture detailed features within a localized window by computing the self-attention within each window. To enhance the global feature capture capability of the model, SW-MSA realizes the information exchange between different windows by shifting the windows so that pixels initially neighboring the boundary but not in the same window can be included in the new window for self-attention calculation. In addition, each Swin Transformer block contains a multilayer perceptron (MLP), which employs residual connection and layer normalization (LN) with the self-attention module and the inputs to enhance training stability. The (S)W-MSA and MLP in the Swin Transformer block are calculated as follows:
\begin{align}
    &Z_{l}^{W-MSA} = W-MSA(LN(Z_{l-1}))+Z_{l-1} \\
    \nonumber \\
    &Z_{l}^{MLP} = MLP(LN(Z_{l}^{W-MSA}))+Z_{l}^{W-MSA} \\
    \nonumber \\
    &Z_{l+1}^{SW-MSA} = SW-MSA(LN(Z_{l}))+Z_{l} \\
    \nonumber \\
    &Z_{l+1}^{MLP} = MLP(LN(Z_{l+1}^{SW-MSA}))+Z_{l+1}^{SW-MSA}
\end{align}
where \(Z_{l}^{(S)W-MSA}\) and \(Z_{l}^{MLP}\) are output features of (S)W-MSA module and MPL module for block \(l\). The design of the W-MSA and SW-MSA allows the Swin Transformer to compute the self-attention in each window, which greatly reduces the computational complexity.
\subsection{Structured Data Feature Extractor}
\begin{figure}[h]
  \centering
  \includegraphics[width=1\linewidth]{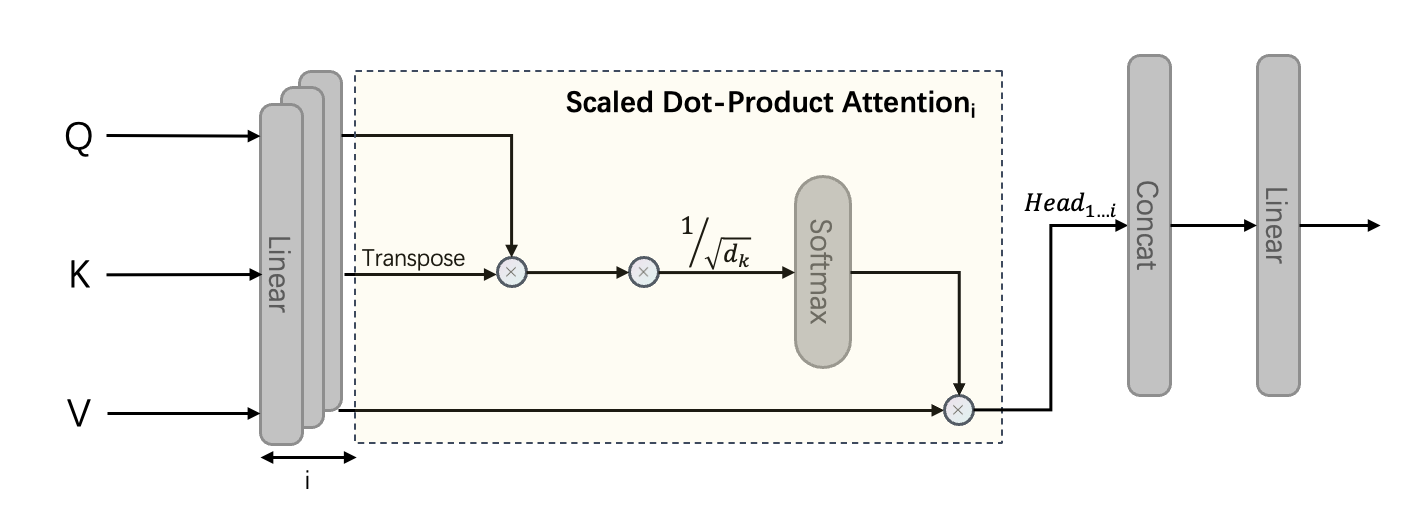}
  \vspace{-0.5cm}
  \caption{Multi-Head Attention architecture.}
  \label{fig:Figure 3}
\end{figure}

Multi-head attention refers to multiple attention modules computing attention independently on different subspaces. This parallel processing (see Figure \ref{fig:Figure 3}) allows the model to capture various information and features from different representation subspaces at different positions. The process of computing the multi-head attention mechanism is as follows: first, the vector matrix \(X\) is obtained, then the vector matrix \(X\) is mapped to different subspaces by the learnable matrices \(W^Q\), \(W^K\), \(W^V\) to obtain matrices Query (\(Q\)), Key (\(K\)), and Value (\(V\)). The \(Q\), \(K\), \(V\) can be written as \(Q = XW^Q\), \(K = XW^K\), \(V = XW^V\), and where $W^Q \in \mathbb{R}^{d_{model} \times d_k}$, $W^K \in \mathbb{R}^{d_{model} \times d_k}$, $W^V \in \mathbb{R}^{d_{model} \times d_v}$. Query matrix (\(Q\)) and the transpose of Key matrix (\(K\)) are used for dot product multiplication, and the result can be obtained as similarity probability after scaling \(\sqrt{d_k}\) times and computed by softmax function, where \(d_k\) is the dimensions of the Key matrix (\(K\)). Finally, the similarity probability is multiplied by the Value matrix (\(V\)) to obtain the attention weight matrix for each attention head. The formula is as follows:
\begin{equation}
\label{eq1}
   Head_{i} = Attention(Q,K,V) = softmax(\frac{QK^{T}}{\sqrt{d_k}})V
\end{equation}
After that, the attention heads will be projected to different subspaces by the learnable weights $W^O \in \mathbb{R}^{{h{d_v}}\times d_{model}}$, where \(d_v\) is the dimensions of the Value matrix (\(V\)) and \(h\) is the number of parallel attention layer, to get the final representation.
\begin{equation}
\label{eq2}
    MHA(Q,K,V) = Concat(Head_{1},\cdots ,Head_{i})W^O
\end{equation}

\vspace{-0.3cm}
Multi-head attention not only allows the networks to learn more information on different subspaces but also increases their stability.

\begin{figure}[h]
  \centering
  \includegraphics[width=0.55\linewidth]{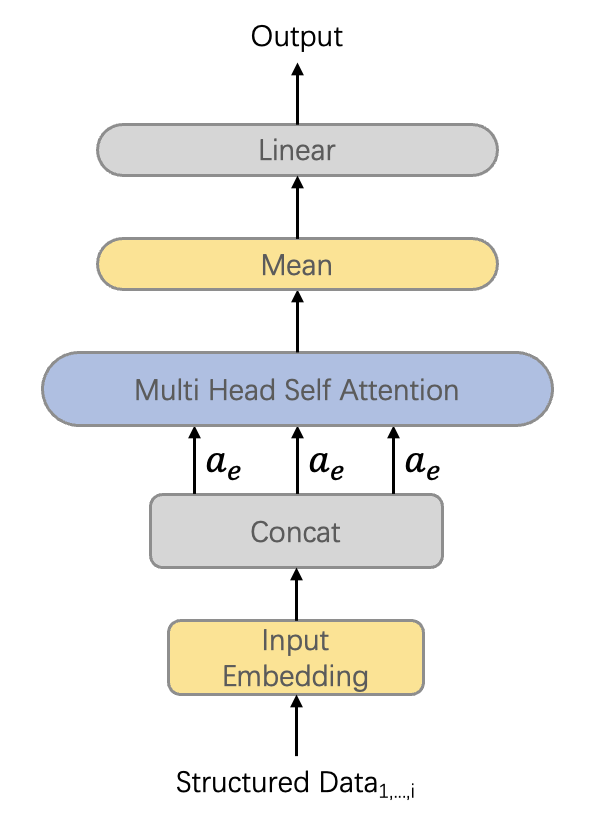}
  \vspace{-0.5cm}
  \caption{Structured Data Feature Extractor architecture.}
  \label{fig:Figure 4}
\end{figure}

Inspired by the multi-head attention mechanism, we apply the multi-head self-attention(MHSA) module in the structured data feature extractor to extract complex relational representations between multiple structured features. Figure \ref{fig:Figure 4} illustrates the structure of the structured data feature extractor. 
First, multiple structured data are transformed into vector representations by embedding layers and concatenated into continuous feature vector (eq. \ref{eq7}). 
\begin{equation}
\label{eq7}
    a_e = Concat(D_1,...,D_i)
\end{equation}
Where \(D_i\) denotes the structured data. Then, continuous feature vector (\(a_e\)) will be input into the multi-head self-attention module as Query matrix (\(Q\)), Key matrix (\(K\)) and Value matrix (\(V\)) respectively. It is worth noting that the difference between the multi-head attention mechanism (MHA) and the multi-head self-attention (MHSA) mechanism is that the multi-head self-attention focuses on dealing with the relationships between elements within a sequence. In contrast, multi-head attention involves dealing with the relationships between two different sequences. After transformation by the multi-head self-attention module, the generated feature vector will contain information about the individual features in the original structured data and the complex relationships between them. Finally, after an averaging operation and a linear layer, the high dimensional feature vectors are downscaled to a dense vector output suitable for the classification task. The averaging formula is shown below:
\vspace{-0.15cm}
\begin{equation}
    output = \frac{1}{n}\sum_{i=1}^{n} Attention(a_{e},a_{e},a_{e})_i
\end{equation}

where \(n\) denotes the length of the dense vector.
\subsection{Multi-Modal Fusion Module}
According to Figure \ref{fig:Figure 1}, the multi-modal fusion strategy used by MMST is decision-layer multi-modal fusion, which weights and fuses the final representations obtained by the image feature extractor and the structured data feature extractor, respectively. The advantages of decision-level multi-modal fusion are scalability and flexibility in handling different modalities. First, decision-layer multi-modal fusion has excellent scalability, allowing different modalities to be added or removed more easily. In addition, the flexibility of the decision-level multi-modal fusion enables the model to choose the most appropriate processing method for each modal data \cite{atrey2010multimodal}. The weighted fusion calculation formula is shown below:
\begin{equation}
    F_{final} = xR + y(1-R)
\end{equation}
Where \(R\) is a preset fusion ratio parameter, \(x\) and \(y\) denote image final features and structured data final features. The fused features are then downscaled by a linear layer to obtain an output format suitable for the classification task.
\section{Case Study}
\subsection{Dataset}

This study uses 2022 Hurricane Ian \cite{ian_data, ian_report} as the case study. This section will introduce the data collection for both street-view imagery and structured data, and discuss the data pre-processing and damage level annotations.
\begin{figure*}[ht!]
    \centering
    \includegraphics[width=\textwidth]{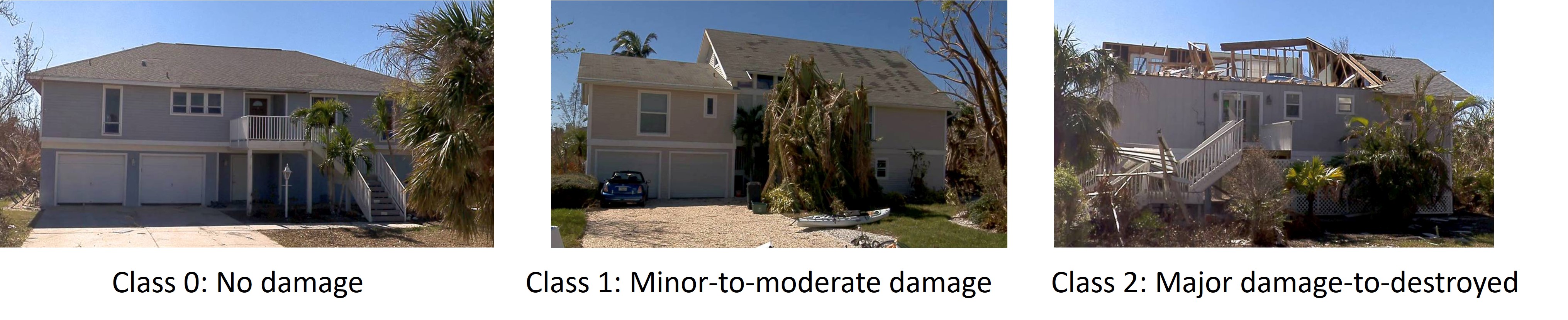}
    \vspace{-0.8cm}
    \caption{An example of building damage level annotation}
    \label{fig:damage_annotation}
\end{figure*}

\vspace{-0.25cm}
\subsubsection{Street-view imagery data}
We used street-view images of buildings affected by Hurricane Ian from two sources: the Structural Extreme Events Reconnaissance (StEER) Hurricane Ian Response\footnote{\url{https://www.steer.network/hurricane-ian}} and our own field investigation. From these efforts, we compiled a building-damage imagery dataset consisting of 2,472 buildings. 
Two annotators were hired to manually assign a damage level to each building, following the guidelines in the StEER's Field Assessment Structural Team (FAST) handbook \cite{kijewski2019field}. This assessment focused solely on the building's exterior. We initially categorized damage into five classes: 0-no damage, 1-Minor damage, 2-Moderate damage, 3-severe damage, and 4-destroyed. However, we found that distinguishing minor or moderate damages (i.e., below 3) was challenging from street-view images and even field investigations. Some subtle signs of damage such as panel cracks were often unidentifiable from images. Therefore, we downscaled the damage levels by merging 1 and 2 as minor-to-moderate damage, 3 and 4 as major damage-to-destroyed, which yields three categories. An example of the damage level-labeled buildings is shown in Fig. \ref{fig:damage_annotation}. We adopted \textit{Krippendorff's alpha} \cite{krippendorff2018content} and correlation coefficient to jointly examine the level of agreement for labels assigned by two annotations. Results showed a 0.86 \textit{Krippendorff's alpha} and 0.97 correlation coefficient, both significant at 0.001 level, indicating consistent annotations. In total, we have 1,781 buildings classified as no damage, 536 buildings as minor-moderate damage, and 151 as major damage to destroyed.
\vspace{-0.25cm}
\subsubsection{Structured data}

Accurately assessing building damage is critical for disaster response and recovery. 
However, many existing models for detecting building damage have poor We included a set of hurricane-damage-related factors as a supplement to facilitate the damage level assessment better. For example, we collected the building age, building value (market value), and the evacuation zone for where the building is located from Lee County Open Data Portal\footnote{\url{https://maps.leegov.com/pages/data}}. We also collected the hurricane track data from National Hurricane Center\footnote{\url{https://www.nhc.noaa.gov/}} and calculated the proximity of each building to the hurricane track using Geographic Information System (GIS) software. We included surface-level wind field data (interpolated) around each building during Hurricane Ian provided by The National Institute of Standards and Technology (NIST) and Applied Research Associates, Inc. (ARA)\footnote{\url{https://www.nhc.noaa.gov/archive/2022/IAN_graphics.php}}.

\vspace{-0.25cm}
\subsection{Implementation}
We built our model using a laptop running Windows 11 with 2.5 GHz 12th generation i9-12900H, 32 GB of RAM, and an NVIDIA 3070ti graphics card. We trained our models in a supercomputer cluster configured with AMD Epy 7742 (Rome) 64-core processors, 2 TB RAM, and NVIDIA A100 80 GB Tensor Core GPUs per node.


\vspace{-0.25cm}
\subsection{Benchmark Models}
\begin{itemize}
\vspace{-0.25cm}
    \item \textbf{Visual Geometry Group(VGG)}: We employed two of the most classical CNN-based models, VGG-16 and VGG-19, as the benchmark models \cite{simonyan2014very, zhai2020damage, xiong2021multiple}. The difference between VGG-16 and VGG-19 is the depth of the network; VGG-16 has 16 layers, including 13 convolutional layers and 3 fully connected layers, while VGG-19 has 19 layers, including 16 convolutional layers and 3 fully connected layers. All of them use 3x3 convolutional filters. The learning rate is 0.0003, the batch size is 32, and they both use the Adam optimizer with a weight decay of 1e-4.
    \vspace{-0.25cm}
    \item \textbf{Residual Network(ResNet)}: We leveraged three mainstream CNN-based models, ResNet-50, ResNet-101, and ResNet-152, as benchmark models \cite{he2016deep,hong2022classification,seydi2022bdd, ci2019assessment}. ResNet-50, ResNet-101, and ResNet-152 are deep residual networks with 48, 99, and 150 convolutional layers, each including 1 maximum pooling layer and 1 average pooling layer. In ResNet-50, the 48 convolutional layers form 16 residual blocks, while in ResNet-101 and ResNet-152, the 99 and 150 convolutional layers form 33 and 50 residual blocks, respectively. These residual blocks help alleviate the problem of gradient dissipation through residual connections. All three models share common hyperparameters: the learning rate is 0.0001, the batch size is 32, and the Adam optimizer has a weight decay 1e-4.
    \vspace{-0.25cm}
    \item \textbf{Vision Transformer(ViT-B16)}: To select the Transformer backbone that performs best on the Hurricane Ian dataset, we utilized ViT-B16 as the benchmark model \cite{yuan2021tokens}. ViT-B16 is an image classification model based on the Transformer architecture, where 16 means the image will be cut into 16*16 pixel patches when it enters the model. The model contains 12 Transformer blocks. The learning rate is 0.0001, the batch size is 32, the number of detectors (head) is 5, and the Adam optimizer has a weight decay 1e-4.
\end{itemize}
\subsection{Performance Evaluation}
A limitation of the Hurricane Ian dataset is that the number of samples in the three classes could be more balanced, with No damage, Minor-to-moderate damage, and Major damage-to-destroyed having sample sizes of 1,786, 536, and 151, respectively. The number of Major damage-to-destroyed is much lower than the other two classes. The model's superior performance in classes with large sample sizes will mask the model's poor performance in classes with small sample sizes. Therefore, we used the Matthews correlation coefficient (MCC) and sample-weighted F1 score (SW-F1 Score) as two evaluation metrics.

By combining true and false positives and negatives, the MCC reflects the proportion of correct and incorrect predictions, providing a more realistic measure of the classifier's effectiveness. The MCC is less sensitive to the class distribution and penalizes inaccurate predictions ranging from -1 to +1, where +1 indicates perfect predictions, 0 is no better than random, and -1 is entirely inconsistent. Thus, the model performance metrics used in this paper are Accuracy, Recall, Precision, SW-F1 Score, and MCC, whose mathematical formulations are provided below. Note that, in this paper, 
we focus more on the model's score performance on the MCC and SW-F1, which are less sensitive to the class imbalance issues \cite{chicco2020advantages}. 
\begin{equation}
    Precision = \frac{TP}{TP+FP}
\end{equation}
\begin{equation}
    Recall = \frac{TP}{TP+FN}
\end{equation}
\begin{equation}
    \textit{SW-F1} = \sum_{i=1}^{N}W_i \times \left ( \frac{2\times Precision\times Recall}{Precision + Recall} \right )_i
\end{equation}
\begin{equation}
\scalebox{0.8}{
$\displaystyle
MCC = \frac{TP\cdot TN - FP\cdot FN}{\sqrt{(TP+FP)\cdot (TP+FN)\cdot (TN+FP)\cdot (TN+FN)}}
$}
\end{equation}
where TP, TN, FP, and FN represent the number of true positives, true negatives, false positives, and false negatives, respectively, \(N\) denotes the number of classes in the dataset, and \(W_i\) denotes the weight of the \(i\) th class. 
\subsection{Hyperparameter Settings}
We used 20\% of the data for testing and the remaining for training and validation. We adopted a stratified sampling approach to ensure equal distribution of the damage levels across three sets. For the image feature extractor, the number of Swin Transformer Blocks in the four stages of Swin-S \cite{liu2021swin} are 2, 2, 18, and 2. For the structured data feature extractor, the number of detectors (heads) is 5. For multi-modal model training, we used a grid search approach to compare the performance of MMST at learning rates of \{0.01, 0.001, 0.0001, 0.00001\} and batch sizes of \{8, 16, 32, and 64\}, respectively. MMST performs best at a learning rate of 0.0001 and a batch size of 32. The step size and gamma of the learning rate descent strategy are set as 5 and 0.85, respectively. The loss function we used is the Focal loss function \cite{Lin_2017_ICCV}, and using the Adam optimizer, weight decay is 1e-4, and epoch is 200. Since there are only 1981 images in the training set, in order to fully utilize this dataset, we adopt a 5-fold cross-validation training strategy.

\vspace{-0.25cm}
\subsection{Comparison of Different Benchmark Models}
We compared the performance of different benchmark models in the Hurricane Ian dataset, which includes VGG-16, VGG-19 \cite{simonyan2014very}, ResNet-50, ResNet-101, ResNet-152 \cite{he2016deep}, ViT-B16 \cite{yuan2021tokens}, Swin-S \cite{liu2021swin}, and MMST. A quantitative comparison of the performance of the different benchmark models for classifying building damages in this data is given in Table \ref{tab:Table 1}. As seen from Table \ref{tab:Table 1}, the Transformer performs much better than traditional convolutional neural networks on this dataset. Among all the benchmark models, the MMST proposed in this study performs the best on this dataset in terms of MCC, SW-F1 score, and accuracy, which are 0.7404, 0.9386, and 92.67\%, respectively, with an improvement of 56.70\%, 5.03\%, and 7.71\% compared to VGG-16, and the highest precision score is achieved by Swin-S. In addition, it is worth noting that the three ResNet benchmark models with different depths all have an SW-F1 score of 0 on this dataset because they fail to predict a major damage-to-destroyed label. This result means that increasing the depth of the ResNet model does not allow the model to learn more helpful information in the image with the major damage-to-destroyed label. Furthermore, among the three ResNet benchmark models, ResNet-101 performs the best, while ResNet-152 and ResNet-50 perform poorly. This result illustrates that the learning ability of ResNet, in this case, is not linearly related to the network depth. 
\begin{table}
\centering
\vspace{-0.4cm}
\caption{Performance of different benchmark models.}
\vspace{1mm} 
\label{tab:Table 1}
\resizebox{\textwidth/2}{!}{ 
\begin{tabular}{l r c c | c c}
\toprule
\textbf{Model}  &\textbf{MCC} & \textbf{SW-F1 Score} & \textbf{Accuracy} & \textbf{Precision} & \textbf{Recall} \\
\midrule
VGG-16 & 0.3201 & 0.8913 & 84.96\% & 0.5512 & 0.4158  \\
VGG-19 & 0.4104 & 0.8950 & 85.94\% & 0.4763 & 0.4501  \\
\midrule
ResNet-50 & 0.2646 & 0 & 84.76\% & 0.4735 & 0.3717  \\
ResNet-101 & 0.3814 & 0 & 86.13\% & 0.4862 & 0.4142 \\
ResNet-152 & 0.3397 & 0 & 85.54\% & 0.4876 & 0.4036 \\
\midrule
ViT-B16 & 0.6604 & 0.9238 & 90.34\% & 0.8036 & 0.6088 \\
Swin-S & 0.6680 & 0.9250 & 91.21\% & 0.8118 & 0.6250 \\
\midrule
\textbf{MMST} & \textbf{0.7404} & \textbf{0.9386} & \textbf{92.67\%} & 0.7987 & 0.6944 \\
\bottomrule
\end{tabular}
}
\end{table}


\vspace{-0.25cm}
\subsection{Ablation Analysis}
We conducted ablation experiments to determine the contribution of each set of structured data in MMST to the classification of building damage. These sets include Evacuation Zone, Building Age, Estimated Value, Distance to the Hurricane Track, and Wind Speed. Ablation analysis involves removing specific components sequentially to observe the changes in model performance and determine the contribution of each element \citep{meyes2019ablation}. Table \ref{tab:Table 2} shows a comparison of the ablation experiments conducted on the Hurricane Ian dataset. We used the MMST containing all structured data (with MHSA) as the benchmark model for these experiments. We analyzed the contribution of each set of structured data to the classification of building damage by observing the performance change of MMST without any set of structured data.
\begin{table}
\vspace{-0.5cm}
\caption{Role of each structured data in MMST.}
\vspace{1mm}
\centering
\label{tab:Table 2}
\resizebox{\textwidth/2}{!}{ 
\begin{tabular}{l r c c | c c c}
\toprule
\textbf{Features index} &\textbf{MCC}  & \textbf{SW-F1 Score} & \textbf{Accuracy} & \textbf{Precision} & \textbf{Recall}  &\textbf{Corr.$^*$} \\
\midrule
w/o Building Age & 0.6731 & 0.9165 & 90.22\% & 0.8193 & \textbf{0.6993}   & \(+\) \\
w/o Building Value & 0.6661 & 0.9242 & 90.42\% & 0.7069 & 0.6462 & \(-\)\\
w/o Wind speed & 0.6746 & 0.9231 & 90.82\% & 0.7946 & 0.6666   & \(+\)\\
w/o Dist. to hurricane track & 0.6845 & 0.9244 & 91.03\% & 0.7983 & 0.6756  & \(-\)\\
w/o Evacuation Zone & 0.6904 & 0.9357 & 91.44\% & \textbf{0.8860} & 0.5874  & \(+\)\\
\midrule
All features (w/o MHSA) & 0.6439 & 0.9254 & 90.21\%  & 0.7580 & 0.5808  \\
All features & \textbf{0.7404} &\textbf{0.9386} & \textbf{92.67\%} & 0.7987 & 0.6944  \\
\bottomrule
\end{tabular}
}
\parbox[t]{0.98\textwidth/2}{\vskip3pt{\footnotesize Corr.$^*$: The direction of correlation between feature values and damage level.}}
\end{table}
When the input structured data does not contain evacuation zone, compared with the benchmark model, the MCC of MMST only decreases by 6.70\%, and the SW-F1 score and accuracy decrease by 0.30\% and 1.23\%, respectively, and when the input structured data does not contain building age, a significant reduction in the performance of MMST occurs. The MCC, SW-F1 score, and accuracy decreased by 9.00\%, 2.34\%, and 2.45\%, respectively. All features (w/o MHSA) denote structured data processing without the multi-head self-attention module. By comparison, it is found that the model performance undergoes an explicit decrease in MCC, SW-F1 score, and accuracy by 13.00\%, 1.40\%, and 2.46\%, respectively.

Table \ref{tab:Table 3} demonstrates the performance impact of different fusion ratios on MMST. The table shows that the model performance gradually increases when the fusion ratio is less than 0.80, reaches its maximum at 0.80, and a decrease in model performance occurs when it is greater than 0.80. When the fusion ratio is 0.80, MMST has the highest accuracy and scores on MCC. In addition, MMST obtained the highest SW-F1 score at a fusion ratio of 0.90.
\begin{table}
\centering
\vspace{-0.4cm}
\caption{Performance of different fusion ratios.}
\vspace{1mm}
\label{tab:Table 3}
\resizebox{\textwidth/2}{!}{ 
\begin{tabular}{l r c c |c c}
\toprule
\textbf{Fusion Ratio} &\textbf{MCC} & \textbf{SW-F1 Score}  & \textbf{Accuracy} & \textbf{Precision} & \textbf{Recall}  \\
\midrule
0.70   & 0.7292 & 0.9376 & 92.46\% & 0.8465 & 0.6666 \\
0.75  & 0.7236 & 0.9355 & 92.26\% & \textbf{0.8463} & 0.6805 \\
\textbf{0.80} & \textbf{0.7404} & 0.9386 & \textbf{92.67\%} & 0.7987 & \textbf{0.6944} \\
0.85  & 0.7063 & 0.9340  & 91.85\% & 0.7667 & 0.6666 \\
0.90   & 0.7036 & \textbf{0.9390}& 91.85\% & 0.7301 & 0.5890\\
\bottomrule
\end{tabular}
}
\end{table}
\begin{figure*}[h]
  \centering
  \includegraphics[width=0.80\linewidth]{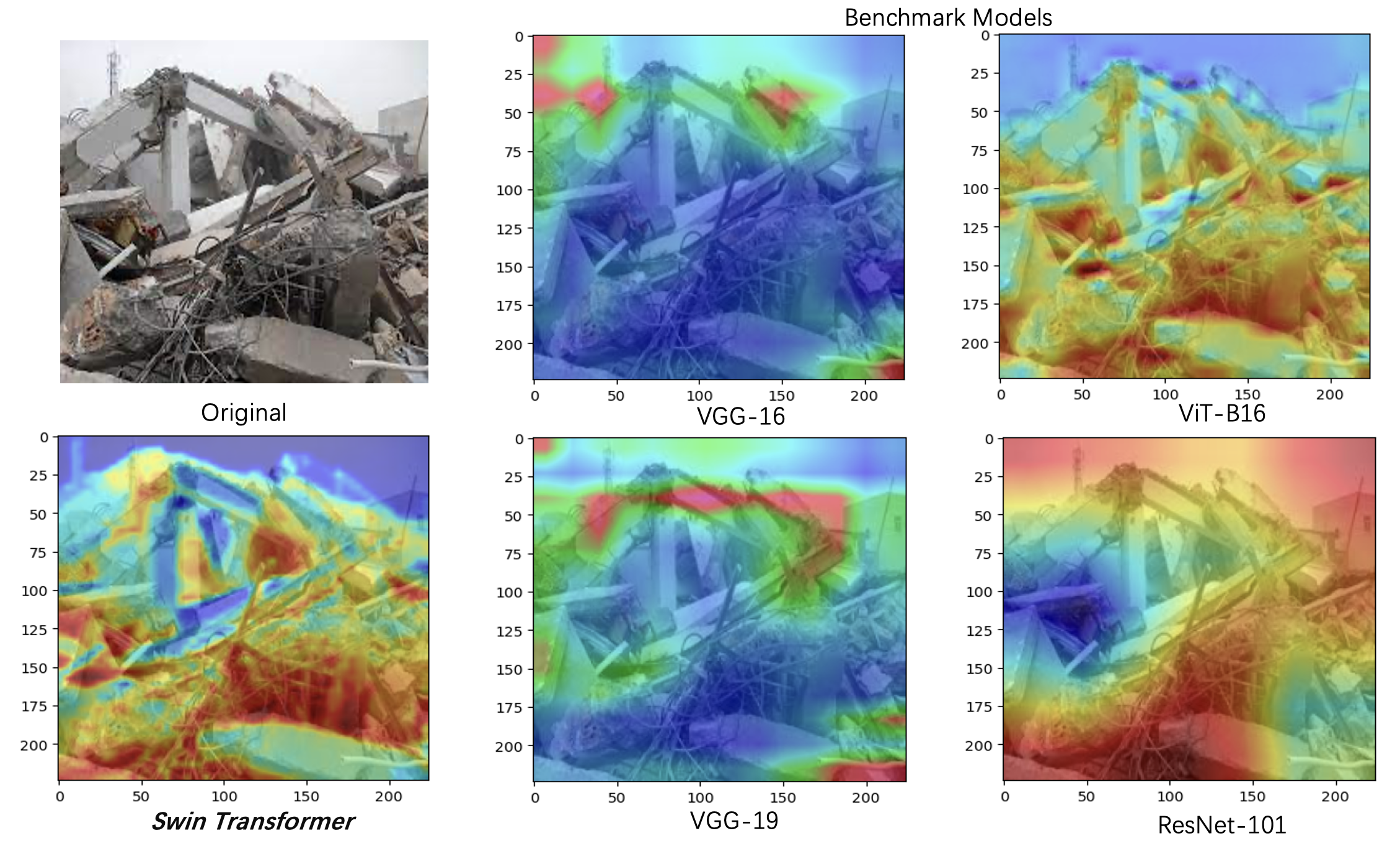}
  \vspace{-0.25cm}
  \caption{Attention visualization with different benchmark models.}
  \label{fig:Figure 5}
\end{figure*}
These results show that the evacuation zone in the Hurricane Ian dataset contributes the least to the decision of MMST in the classification of building damages. In contrast, the building age contributes the most. In addition, including the multi-head self-attention module while processing structured data can effectively improve the performance and robustness of the model. This result is because in the process of multi-head self-attention computation, multiple sets of structured data features contained in the dense vectors will be given different expression weights, and the more critical structured data features will play a more important role in the classification decision. Finally, by comparing different fusion ratios, we learned that the performance of MMST is highest when the fusion ratio is 0.80.
\section{Discussion}
First, incorporating multi-modal data into deep learning models can improve the model's prediction accuracy. Our empirical study (as shown in Table \ref{tab:Table 2}) showed that in addition to imagery data, structured data such as building characteristics and wind speed, also played significant roles in determining building damage levels. The results are consistent with previous studies \cite{cheng2021deep, fronstin1994determinants, egnew2018linking, xian2015storm, jain2009statistical,kim2016predicting}. These structured data contain valuable information that a single image cannot capture. Supplementing the imagery data with these factors brings the model's predictions closer to empirical data and observations in the real world. Also, including these structured data aligns with our domain knowledge. For example, this study showed that among all structured data, building age is a critical predictor for forecasting hurricane damage \cite{kim2016predicting}. The older the building is, the more susceptible it is to damage. In recent years, there has been a recognition that integrating domain knowledge into a deep learning model can enhance the model's performance \cite{borghesi2020improving}. Our study further confirms this point by contributing insights into the post-hurricane building damage assessment domain.

Second, the Transformer can focus more accurately and precisely on building damage information and patterns than traditional CNN-based architecture models. Traditional CNN-based models (e.g., VGG and ResNet) can only detect local features using a fixed-size convolutional kernel, resulting in models that only partially understand building damage \cite{sriwong2021study}. This structural limitation also tends to confuse noise information with important building damage information during feature extraction, making it difficult to improve the model's accuracy \cite{arkin2023survey}. On the other hand, Transformer integrates an attention module that dynamically focuses attention on the most salient parts of the image, especially the critical damaged areas in the buildings \cite{kaur2023large,chen2022dual}. As a result, Transformers improve the accuracy and stability of classifying building damage patterns. Figure \ref{fig:Figure 5} illustrates the regions within an image that different backbone networks focus on when making classification decisions, where the closer to the red color represents that this part has a higher level of attention from the model, and vice versa. This figure visually demonstrates the advantages of the Transformer in building damage assessment tasks.


\section{Conclusion}
In this work, we propose a novel multi-modal classification method called MMST to address the problem of insufficient building damage information in existing post-hurricane building damage classification models with single-modal data. MMST was trained and evaluated from the Hurricane Ian dataset, showing that the MMST significantly outperforms conventional CNN models. Compared to VGG-16, MMST improves 56.70\%, 5.03\%, and 7.71\% in MCC, SW-F1 score, and accuracy, respectively. Furthermore, by incorporating structured data, MMST's MCC, SW-F1 score, and accuracy improved by 9.70\%, 1.44\%, and 1.38\%, respectively, over the Swin Transformer with the single imagery modality. 

Although MMST has outperformed CNN-based models and Transformer that use a single imagery modality on the Hurricane Ian dataset, MMST still has some limitations, such as choosing a better Transformer as the backbone of the image feature extractor, incorporating more diverse modality data and considering different multi-modal data fusion strategies. In the future, we will enhance MMST by testing more advanced Transformer models as the backbone, integrating a more comprehensive range of modality data, and trying different fusion strategies. Meanwhile, to help local governments, insurance companies, and emergency management make more accurate building damage predictions, we will extend the application of MMST to larger-scale datasets and disaster-related tasks, such as flood prediction or earthquake damage assessment, to evaluate its adaptability and generalization in different disastrous scenarios.
\bibliography{main}
\bibliographystyle{icml2021}

\end{document}